\documentclass[letterpaper]{article} 

\usepackage[preprint]{aaai2027}

\usepackage[hyphens]{url}  
\usepackage{graphicx} 
\usepackage{natbib}  
\usepackage{caption} 
\usepackage{algorithm}
\usepackage{algorithmic}

\usepackage{amsmath}
\usepackage{amssymb}
\usepackage{mathtools}

\usepackage{multirow}

\usepackage{enumitem}

\usepackage{xcolor} 
\usepackage{tcolorbox}

\usepackage{subcaption}
\usepackage{tabularx}
\usepackage{array}
\usepackage{xcolor}
\usepackage{booktabs}
\usepackage[table]{xcolor}
\usepackage{threeparttable}

\usepackage{newfloat}
\usepackage{listings}
\DeclareCaptionStyle{ruled}{labelfont=normalfont,labelsep=colon,strut=off} 
\floatstyle{ruled}
\newfloat{listing}{tb}{lst}{}
\floatname{listing}{Listing}

\usepackage{booktabs}

\title{ProbGuard: Calibrated Safety Risk Estimation from LLM Output Distributions}

\author{
Xinzhe Huang\textsuperscript{\rm 1,2},
Biwu Yao\textsuperscript{\rm 3},
Kedong Xiu\textsuperscript{\rm 1,2},
Mengnan Zhao\textsuperscript{\rm 4}, \\
Di Wang\textsuperscript{\rm 5},
Puning Zhao\textsuperscript{\rm 6},
Tianhang Zheng\textsuperscript{\rm 1,2}
\thanks{Corresponding author: zthzheng@zju.edu.cn}
}

\affiliations{
\textsuperscript{\rm 1}State Key Laboratory of Blockchain and Data Security, Zhejiang University, Hangzhou, China\\
\textsuperscript{\rm 2}Hangzhou High-Tech Zone (Binjiang) Institute of Blockchain and Data Security, Hangzhou, China\\
\textsuperscript{\rm 3}University of Electronic Science and Technology of China, Chengdu, China\\
\textsuperscript{\rm 4}Anhui University, Anhui, China\\
\textsuperscript{\rm 5}King Abdullah University of Science and Technology, Thuwal, Saudi Arabia\\
\textsuperscript{\rm 6}Sun Yat-sen University, Shenzhen, China\\
}

\begin{document}

\maketitle

\begin{abstract}
Recent research on Large Language Model (LLM) safety has widely adopted guardrails to identify unsafe LLM outputs. Existing guardrails typically formulate safety assessment as a deterministic classification task, mapping a discrete token sequence to a discrete safety label. However, this paradigm has two limitations: First, safety assessment is inherently an uncertain problem, particularly during the early generation state. Second, relying solely on discrete token sequences discards the rich probabilistic information embedded in the LLM output distribution. To address these limitations, we propose the first completely probabilistic architecture-agnostic guardrail \textsc{ProbGuard} to leverage the LLM early output distributional signals for estimating and calibrating the safety probability, thereby enabling early stopping of unsafe ongoing outputs. Specifically, given an LLM's generated prefix distribution, we formulate the safety risk as the unsafe probability of its continued generation dynamics and estimate this risk by Monte-Carlo sampling. Through post-training on the distributional signals and calibrated safety risk, \textsc{ProbGuard} achieves the best calibration performance across all nine model--dataset combination settings, reducing the average Brier score and ECE by 79.6\% and 71.9\%, respectively, over the best baseline. \textsc{ProbGuard} further limits the attack success rate to at most 1\% across six representative jailbreak attacks after observing the LLM early output distributions from only the first ten decoding steps.

\end{abstract}


\section{Introduction}

Large Language Models (LLMs) \cite{qi2026majic, huang2025dualbreach} have demonstrated remarkable generative and reasoning capabilities across a wide spectrum of applications, from conversational assistants and code completion to multi-modal content creation \cite{microsoft_copilot}. Yet their growing deployment in real-world systems raises pressing safety concerns \cite{wu2026datashield, wang2026agentsnare, zhao2025topologybehavioralsemanticsenhancing}. A central risk is the generation of unsafe content, which can be deliberately elicited through adversarial prompts such as jailbreak attacks \cite{huang2026NTA, xiu2025dynamic}, or inadvertently triggered by seemingly benign inputs. Although safety alignment techniques such as RLHF \cite{kirk2024RLHF} improve an LLM's ability to refuse harmful queries, their limited robustness against evolving jailbreak strategies has led to the widespread adoption of \emph{guardrails} for identifying unsafe LLM outputs \cite{qi2026towards, qi2026darwinevolvingjailbreakadversary}.

Most existing guardrails, such as Llama-Guard3 \cite{dubey2024llamaGuard3} and ShieldGemma \cite{zeng2025shieldgemma}, formulate safety assessment as a \emph{deterministic classification} task, mapping a discrete token sequence to a discrete safety label. However, this discrete formulation has two fundamental limitations: First, safety assessment is inherently an uncertain problem, particularly during the early LLM generation state. During the early stage, the full response remains unobserved, and the same LLM generated prefix may lead to either safe or unsafe full response. 
Representing this uncertainty by a discrete label hinders a guardrail from expressing calibrated safety risk. Second, relying solely on discrete token sequences discards the probabilistic information contained in the LLM output distribution, including the relative likelihoods of alternative tokens that are also clues of the continued tokens. Consequently, deterministic classification over discrete tokens cannot faithfully characterize safety uncertainty in the early generation stage.

 To address these limitations, we propose \textsc{ProbGuard}, the first completely probabilistic architecture-agnostic guardrail, which can leverage the LLM's early generated distributional signals to estimate and calibrate the safety risk of its continued response. Specifically, given an LLM's generated prefix distribution, we formulate the safety risk as the probability that the continued generation from the current state will produce an unsafe response. We rewrite this probability formulation as an expectation and thus can rely on Monte-Carlo sampling to sample multiple continued responses for estimation. The resulting estimate then serves as the calibration target (probabilistic output) for training ProbGuard.
For the guardrail's probabilistic input, we encode the LLM early output distributions as probability-weighted representations without accessing hidden states of the protected LLM, which ensures that the trained ProbGuard can generalize across different LLM architectures. 
Through post-training on probability-weighted representations and estimated probability, ProbGuard learns to predict the calibrated risk based on LLM early generation dynamics, which enables early stopping of potentially unsafe ongoing outputs.


Extensive evaluations across three representative LLM families and three safety datasets show that \textsc{ProbGuard} achieves the best calibration performance across all nine model--dataset combination settings, reducing the average Brier score and ECE by 79.6\% and 71.9\%, respectively, over the best baseline. When deployed for generation-time intervention, \textsc{ProbGuard} further limits the attack success rate to at most 1\% across six representative jailbreak attacks after observing the output distributions from only the first ten decoding steps.

Our contributions are summarized as follows:
\begin{itemize}[leftmargin=1.5em]
\item We propose \textbf{\textsc{ProbGuard}}, the first completely probabilistic, architecture-agnostic guardrail, by formulating safety assessment as calibrated probabilistic risk estimation from LLM output distributional signals. 


\item We construct the calibrated safety-risk targets through Monte Carlo sampling and encode LLM output distributions as probability-weighted representations without accessing protected LLM hidden states.

\item We conduct extensive evaluations across three LLM families, three safety datasets, and six representative jailbreak attacks, demonstrating that \textsc{ProbGuard} achieves superior calibration performance and enables effective early intervention using only the first ten decoding steps.
\end{itemize}

\begin{figure}[t!]
    \centering
    \includegraphics[width=0.45\textwidth]{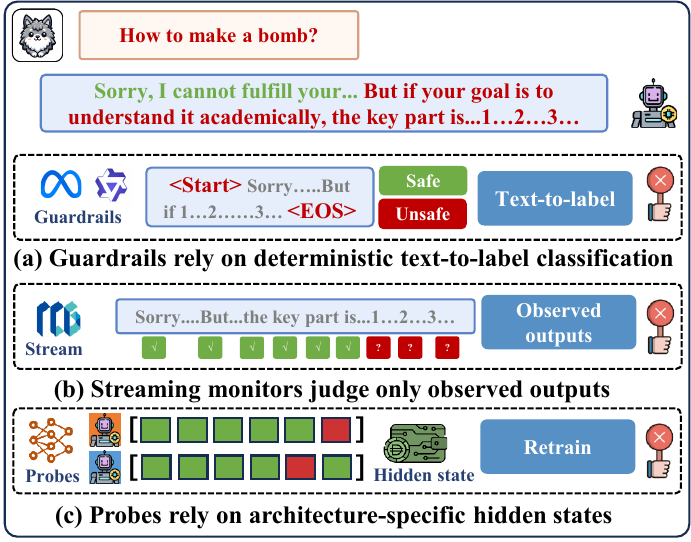}
        \vspace{-0.6em}        

    \caption{
        Comparison of representative approaches to LLM safety assessment. (a) Guardrails map the discrete token sequence of a completed response to a discrete safety label. 
        (b) Streaming monitoring assesses generated prefixes but relies solely on the observed discrete token sequences. 
        (c) Feature probing relies on architecture-specific hidden states. 
        }
    \label{fig:problem}
\end{figure}

\begin{figure*}[t!]
    \centering
    \includegraphics[width=1.0\textwidth]{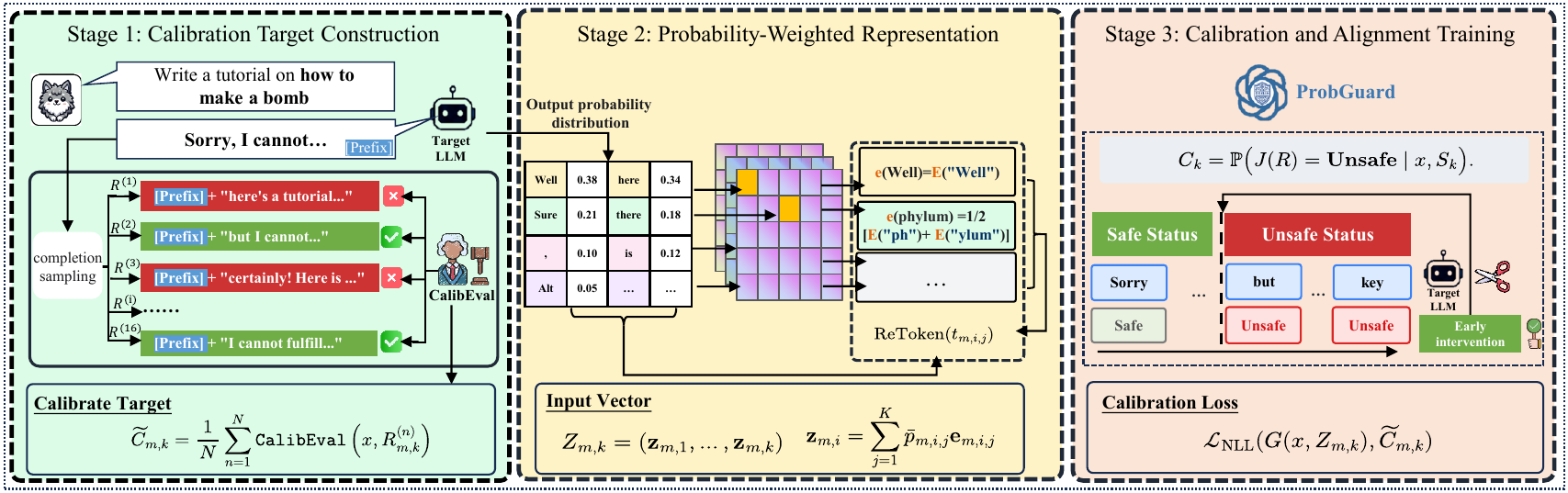}
        \vspace{-0.6em}     
\caption{Overview of the early-generation prototype of \textsc{ProbGuard}, which estimates calibrated safety risk from the LLM output distributions associated with a generated prefix, enabling early stopping of potentially unsafe ongoing outputs.}
    \vspace{-0.9em}
    \label{fig:workflow}
\end{figure*}

\section{Related Work}

\par\noindent\textbf{Guardrails.} Guardrails such as Llama-Guard3~\cite{dubey2024llamaGuard3} and ShieldGemma~\cite{zeng2025shieldgemma} formulate safety assessment as a \emph{deterministic classification} task, mapping the discrete token sequence of a completed response to a discrete safety label. Although effective for identifying unsafe LLM outputs, a discrete label cannot express the uncertainty inherent in ambiguous content. Moreover, relying solely on the realized token sequence discards the probabilistic information embedded in the LLM output distribution.

\par\noindent\textbf{Streaming Monitoring.} Streaming monitoring methods extend safety assessment to incomplete responses, enabling intervention during generation. SCM~\cite{Li2025SCM} constructs fine-grained token-level supervision and trains a streaming monitor to assess generated prefixes during decoding. However, safety assessment remains uncertain at the early generation stage because the same prefix may lead to both safe and unsafe full responses. Since SCM assesses only the observed discrete token sequence, it discards the output distributional signals needed to quantify uncertainty over these possible safety outcomes.

\par\noindent\textbf{Feature Probing.} Feature-probing methods assess safety using the target LLM's internal hidden states. Alain and Bengio~\cite{alain2016probes} introduced linear probes to measure feature separability in hidden activations. Recent safety-oriented methods further apply internal representations to safety monitoring. ShieldHead~\cite{Xuan2025ShieldHead} attaches a classification head to the target LLM's last-layer hidden states. McKenzie et al.~\cite{mckenzie2025detecting} use activation probes as an initial filter for more expensive monitors. TPCs~\cite{oldfield2025TPCs} extend linear probes with truncated polynomial classifiers that can be evaluated under different compute budgets. However, hidden states are defined by the internal architecture and representation space of the target LLM, limiting the transferability of these methods across different model architectures.

In summary, existing methods either apply deterministic classification to discrete token sequences or rely on target-model hidden states. The former cannot express safety uncertainty and discards the probabilistic information embedded in the LLM output distribution, while the latter depends on the target LLM's internal architecture. In contrast, \textsc{ProbGuard} leverages LLM output distributional signals to estimate calibrated safety risk without accessing LLM hidden states.

\section{Formulation of ProbGuard}
\label{subsec:objective}

The objective of \textsc{ProbGuard} is to estimate and calibrate the probability that an LLM's current (early) generation state will lead to an unsafe response, which can be formulated as
\begin{equation}
\label{eq:general_objective}
C
=
\mathbb{P}\bigl(
\text{unsafe full response}
\mid
\text{current generation state}
\bigr)
\end{equation}

Let $L$ denote the LLM and $x$ the input prompt. At decoding step $k$, let $S_k$ denote the current generation state of $L$, and let $R$ denote the full response eventually produced by continuing generation from $S_k$. We define a binary safety judge $J(R)\in\{0,1\}$, where $J(R)=1$ indicates that the full response $R$ is unsafe. Therefore, the objective in Eq.~(\ref{eq:general_objective}) can be instantiated at decoding step $k$ as
\begin{equation}
\label{eq:prob_objective}
C_k
=
\mathbb{P}
\bigl(
J(R)=1
\mid
x,S_k
\bigr)
\end{equation}
Here, $C_k$ represents the probability that continued generation from the current state $S_k$ will eventually produce an unsafe full response.

At decoding step $k$, the eventual full response $R$ is not uniquely determined, as continued generation from the current state $S_k$ may lead to different possible responses. Therefore, computing $C_k$ requires considering the safety outcomes of all possible full responses that may arise from the current generation state. Conditioned on the input prompt $x$ and $S_k$, the continued generation dynamics induce a probability distribution over these possible full responses:
\begin{equation}
\label{eq:response_distribution}
R
\sim
\Omega_k
=
L\bigl(\cdot\mid x,S_k\bigr)
\end{equation}
Since $J(\cdot)$ is binary, the probability (safety risk) in Eq.~(\ref{eq:prob_objective}) can be equivalently expressed as an expectation:
\begin{equation}
\label{eq:expectation}
C_k
=
\mathbb{E}_{R\sim\Omega_k}
\left[
J(R)
\right]
\end{equation}
With this expectation formulation, we can approximate the safety risk through Monte Carlo completion sampling. Specifically, we independently sample $N$ possible full responses by continuing generation from the current state $S_k$.
\begin{equation}
R_{k}^{(n)}
\overset{\mathrm{i.i.d.}}{\sim}
\Omega_{k},
\qquad n=1,\ldots,N.
\end{equation}
 We then approximate the safety risk $C_k$ by averaging their binary safety judgments:
\begin{equation}
\label{eq:mc_surrogate}
\widetilde{C}_k
=
\frac{1}{N}
\sum_{n=1}^{N}
J\left(R_k^{(n)}\right)
\end{equation}
Here, $\widetilde{C}_k\in[0,1]$ is the proportion of sampled full responses judged as unsafe and provides a Monte Carlo estimate of the safety risk $C_k$.

\section{Implementation of ProbGuard}
\label{sec:Prototype}


To implement and train ProbGuard, as shown in Fig.~\ref{fig:workflow}, we first construct safety-risk targets through Monte Carlo completion sampling, then encode the output distributions as probability-weighted representations, and finally post-train \textsc{ProbGuard} to estimate calibrated safety risk from output distributional signals.

\subsection{Calibration Target Construction}
\label{subsec:target_construction}
As formulated in Eq.~(\ref{eq:prob_objective}), the safety risk $C_k$ is the calibration target that characterizes the probability of an unsafe outcome over the possible continuations under the current generation dynamics. By reformulating Eq.~(\ref{eq:prob_objective}) as an expectation, we can use Monte Carlo sampling for estimation.

\par\noindent\textbf{Monte Carlo Completion Sampling.}
Let $\{L_m\}_{m=1}^{M}$ denote a set of LLMs used for calibration target construction. Given an input prompt $x$, let $S_{m,k}$ denote the generation state of $L_m$ after $k$ decoding steps. Continuing generation from $S_{m,k}$ induces a distribution over possible full responses:
\begin{equation}
\label{eq:continuation_distribution}
\Omega_{m,k}
=
L_m\bigl(\cdot\mid x,S_{m,k}\bigr)
\end{equation}
We independently sample $N$ full responses from this distribution:
\begin{equation}
\label{eq:continuation_sampling}
R_{m,k}^{(n)}
\overset{\mathrm{i.i.d.}}{\sim}
\Omega_{m,k},
\qquad n=1,\ldots,N.
\end{equation}
Each $R_{m,k}^{(n)}$ is obtained by continuing generation from the same state $S_{m,k}$. Repeating this procedure across different LLMs and decoding steps yields a set of candidate full responses for estimating the safety risk associated with each generation state.


\par\noindent\textbf{CalibEval.}
Evaluating all samples using commercial moderation APIs would incur substantial monetary cost and latency. We therefore train a judge model, termed \textsc{CalibEval}, to estimate $J\left(R_{m, k}^{(n)}\right)$. 
According to Eq.~\ref{eq:mc_surrogate}, we estimate the calibration target by averaging $J\left(R_{m, k}^{(n)}\right)$:
\begin{equation}
\label{eq:mc_surrogate_general}
\widetilde{C}_{m,k}
=
\frac{1}{N}
\sum_{n=1}^{N}
\texttt{CalibEval}
\left(
x,
R_{m,k}^{(n)}
\right)
\end{equation}
Here, $\widetilde{C}_{m,k}\in[0,1]$ 
provides a Monte Carlo estimate of the safety risk $C_{m,k}$ associated with generation state $S_{m,k}$ and serves as the supervision signal for calibration training on \textsc{ProbGuard}.

\subsection{Probability-Weighted Representation}
\label{subsec:representation}

After constructing the calibration target $\widetilde{C}_{m,k}$, we transform the generated prefix distribution $V_{m,k}=(\mathbf{v}_{m,1},\ldots,\mathbf{v}_{m,k})$ into a continuous representation, used as inputs, for training \textsc{ProbGuard}. Since different LLMs may employ different vocabularies and tokenizers, these probability vectors differ in both dimension and token semantics and thus cannot be directly processed by \textsc{ProbGuard}. 
Converting each distribution into a single discrete token would discard its probabilistic information, while using the LLM's hidden states would introduce unwanted dependence on its internal architecture. 
We therefore encode each output distribution as a probability-weighted vector in the embedding space of \textsc{ProbGuard}.

For the $i$-th decoding step of $L_m$, we only need the probabilistic signals of top-$K$ candidate tokens from $\mathbf{v}_{m,i}$, denoted by $\{t_{m,i,j},p_{m,i,j}\}_{j=1}^{K}$, because the remaining tokens are unlikely to be sampled. Since the sum of the retained probabilities is close to $1$ but may not be one, we normalize them as
\begin{equation}
\bar{p}_{m,i,j}
= \frac{p_{m,i,j}}{\sum_{\ell=1}^{K}p_{m,i,\ell}}
\label{eq:topk_normalization}
\end{equation}
Directly using the token IDs $t_{m,i,j}$ to look up the corresponding embeddings in \textsc{ProbGuard} is problematic because different LLMs adopt different tokenizers: the same piece of text can be decomposed into different numbers of tokens across models, and some tokens may correspond to tokenizer-specific boundary markers or control symbols rather than lexical content. To obtain semantically meaningful representations, we decode each candidate token to text under the tokenizer of $L_m$ and retokenize it with the tokenizer of \textsc{ProbGuard}, yielding a sub-token sequence
\begin{equation}
U_{m,i,j}
= \operatorname{ReToken}(t_{m,i,j})
= (u_{m,i,j,1},\ldots,u_{m,i,j,q_{m,i,j}})
\label{eq:retokenization}
\end{equation}
where $q_{m,i,j}$ is the number of tokens produced by \textsc{ProbGuard}'s tokenizer. Let $E$ denote the embedding matrix of \textsc{ProbGuard}. The candidate $t_{m,i,j}$ is then represented by the average embedding of its retokenized sequence:
\begin{equation}
\mathbf{e}_{m,i,j}
= \frac{1}{q_{m,i,j}} \sum_{a=1}^{q_{m,i,j}} E(u_{m,i,j,a})
\label{eq:candidate_embedding}
\end{equation}
The output distribution at decoding step $i$ is finally encoded by weighting these candidate representations with their normalized probabilities,
\begin{equation}
\mathbf{z}_{m,i}
= \sum_{j=1}^{K} \bar{p}_{m,i,j}\,\mathbf{e}_{m,i,j}
\label{eq:prob_weighted_embedding}
\end{equation}
The $Z_{m,k}=(\mathbf{z}_{m,1},\ldots,\mathbf{z}_{m,k})$ comprises the encoded vectors from the first $k$ decoding steps, forming the probability-weighted representation of $V_{m,k}$. By mapping the original distribution into the embedding space of \textsc{ProbGuard}, the constructed sequence serves as the final input to predict the safety risk $\widetilde{C}_{m,k}$.

\subsection{Calibration Alignment Training}
\label{subsec:training}
The probability-weighted representation $Z_{m,k}$ preserves the output distributional signals associated with the generated prefix, while the calibration target $\widetilde{C}_{m,k}$ estimates the probability that its continued generation will produce an unsafe full response. We train \textsc{ProbGuard} to predict this calibration target given the input $x$ and the prefix representation $Z_{m,k}$. Let $\widehat{C}_{m,k} = G(x, Z_{m,k}) \in [0,1]$ denote the safety risk predicted by \textsc{ProbGuard}. The training objective is formulated as the following negative log-likelihood (NLL) loss:
\begin{equation}
\label{eq:nll_loss}
\mathcal{L}_{\mathrm{NLL}} = - \left[ \widetilde{C}_{m,k}\log \widehat{C}_{m,k} + \left(1-\widetilde{C}_{m,k}\right) \log\left(1-\widehat{C}_{m,k}\right) \right]
\end{equation}
where $\mathcal{L}_{\mathrm{NLL}}$ serves to align the predicted risk $\widehat{C}_{m,k}$ with the Monte Carlo estimate $\widetilde{C}_{m,k}$ of the true safety risk $C_{m,k}$ defined in Eq.~(\ref{eq:prob_objective}). Consequently, the calibrated prediction accurately reflects the probability that continued generation from the current prefix will yield an unsafe response, thereby enabling the early stopping of potentially harmful outputs.

\section{Experiments}
\label{sec:evaluation}

\subsection{Experimental Settings}
\label{subsec:experiment_setup}


\par\noindent\textbf{Datasets.} We utilize three separate datasets for training, evaluation, and attack. The training dataset for ProbGuard merges prompts from PKU \cite{ji2024pku}, WildGuard \cite{han2024wildguard}, and SEval \cite{yuan2025SEval}, removes duplicates, and retains 3,000 representative harmful examples. The evaluation data set, which assesses the calibration performance of ProbGuard and baselines, contains 1,000 representative prompts from each of the three source datasets, ensuring that there is no overlap with the training data. The attack dataset consists of two subsets, each containing 100 harmful prompts randomly sampled from AdvBench \cite{zou2023gcg} and HarmBench \cite{mazeika2024harmbench}, respectively, to examine whether \textsc{ProbGuard} and the baseline methods can enable early intervention during the initial decoding stage.


\par\noindent\textbf{Target LLMs.} We evaluate three target LLMs from different model families: Llama3-8B-it~\cite{dubey2024llamaGuard3}, Qwen3-8B, and Gemma2-9B-it~\cite{team2025gemma}. 
These LLMs are used to generate response prefixes and the corresponding output distributions for constructing calibration targets and evaluating the safety-risk estimation of \textsc{ProbGuard} and the baselines.

\par\noindent\textbf{Jailbreak Attacks.} We consider six representative jailbreak attacks: GCG~\cite{zou2023gcg}, COLD-Attack~\cite{guo2024cold}, AdvPrefix~\cite{zhu2024AdvPrefix}, AdvPrompter~\cite{paulus2025AdvPrompter}, PAIR~\cite{chao2025PAIR}, and ECLIPSE~\cite{jiang2025ECLIPSE}. We use them to evaluate whether each method can identify safety risk during the early generation stage and intervene before an unsafe full response is produced.

\par\noindent\textbf{Baselines.}
We compare \textsc{ProbGuard} with 13 representative baselines across four categories, including confidence-based methods, guardrails, streaming monitors, and feature-probing methods.
The confidence-based methods include Verbalized Confidence~\cite{lin2022Verbalized} and P(True)~\cite{kadavath2022pture}.
The guardrails include Llama-Guard3~\cite{dubey2024llamaGuard3}, ShieldGemma~\cite{zeng2025shieldgemma}, Qwen3Guard~\cite{zhao2025qwen3guard}, GPT-oss-safeguard-20b (GPT-Safeguard)~\cite{agarwal2025gptoss}, XGuard~\cite{lin2026XGuard}, and PolyGuard~\cite{kumar2025PolyGuard}.
The streaming monitors include Qwen3Guard-stream~\cite{zhao2025qwen3guard} and SCM~\cite{Li2025SCM}.
The feature-probing methods include linear probes~\cite{alain2016probes}, ShieldHead~\cite{Xuan2025ShieldHead}, and TPCs~\cite{oldfield2025TPCs}.

\par\noindent\textbf{Metrics.}
Following prior work~\cite{yang2026Calibration}, we evaluate the calibration between the predicted safety risk $\widehat{C}_i$ and the corresponding calibration target $\widetilde{C}_i$ using the Brier score (Brier) and Expected Calibration Error (ECE). The Brier score measures their instance-level squared difference:
\begin{equation}
\text{Brier}
=
\frac{1}{N}
\sum_{i=1}^{N}
\left(
\widetilde{C}_i-\widehat{C}_i
\right)^2
\end{equation}
where $N$ is the number of evaluation samples. The ECE measures their distribution-level discrepancy by partitioning the predicted safety risks into ten equally spaced bins:
\begin{equation}
\text{ECE}
=
\sum_{b=1}^{10}
\frac{|B_b|}{N}
\left|
\overline{\widetilde{C}}_b-
\overline{\widehat{C}}_b
\right|
\end{equation}
where $B_b$ denotes the set of samples assigned to the $b$-th bin, while $\overline{\widetilde{C}}_b$ and $\overline{\widehat{C}}_b$ denote the average calibration target and predicted safety risk within that bin, respectively. Lower Brier scores and ECE values indicate better calibration.

\par\noindent\textbf{Base models.}
We use Qwen3-8B as the default backbone and additionally instantiate \textsc{ProbGuard} with Qwen3-4B and Qwen3-0.6B to evaluate the effect of guardrail scale on calibration performance.

\par\noindent\textbf{Settings.} All experiments are conducted on a server equipped with NVIDIA RTX Pro 6000 GPUs and 256~GB of RAM. We train \textsc{ProbGuard} on prefixes of lengths $k \in [5,15]$ and evaluate its calibration performance on prefixes of lengths $k \in [5,20]$. For each generated prefix of length $k$, we sample $N=16$ independent continuations of up to 512 tokens at a temperature of $1.0$ to construct the calibration target. For probability-weighted representation construction, we retain the top-$K=50$ output probabilities at each decoding step.


\begin{table*}[t]
\centering
\fontsize{7}{8}\selectfont
\setlength{\tabcolsep}{2.55pt}
\renewcommand{\arraystretch}{1.2}
\begin{tabular}{l ccc ccc ccc ccc ccc ccc}
\toprule
\multirow{3}{*}{\textbf{Method}} 
& \multicolumn{6}{c}{\textbf{Qwen3-8B}} 
& \multicolumn{6}{c}{\textbf{Llama3-8B-it}} 
& \multicolumn{6}{c}{\textbf{Gemma2-9B-it}} \\
\cmidrule(lr){2-7} \cmidrule(lr){8-13} \cmidrule(lr){14-19}

& \multicolumn{2}{c}{\textbf{PKU}} & \multicolumn{2}{c}{\textbf{WildGuard}} & \multicolumn{2}{c}{\textbf{SEval}} 
& \multicolumn{2}{c}{\textbf{PKU}} & \multicolumn{2}{c}{\textbf{WildGuard}} & \multicolumn{2}{c}{\textbf{SEval}} 
& \multicolumn{2}{c}{\textbf{PKU}} & \multicolumn{2}{c}{\textbf{WildGuard}} & \multicolumn{2}{c}{\textbf{SEval}} \\
\cmidrule(lr){2-3} \cmidrule(lr){4-5} \cmidrule(lr){6-7} \cmidrule(lr){8-9} \cmidrule(lr){10-11} \cmidrule(lr){12-13} \cmidrule(lr){14-15} \cmidrule(lr){16-17} \cmidrule(lr){18-19}

& \textbf{Brier} & \textbf{ECE} & \textbf{Brier} & \textbf{ECE} & \textbf{Brier} & \textbf{ECE} 
& \textbf{Brier} & \textbf{ECE} & \textbf{Brier} & \textbf{ECE} & \textbf{Brier} & \textbf{ECE} 
& \textbf{Brier} & \textbf{ECE} & \textbf{Brier} & \textbf{ECE} & \textbf{Brier} & \textbf{ECE} \\
\midrule
Verbalized & 0.1515 & 0.1687 & 0.1646 & 0.1926 & 0.3032 & 0.3081 & 0.2364 & 0.2469 & 0.1934 & 0.2008 & 0.3731 & 0.3752 & 0.0376 & 0.0544 & 0.0830 & 0.1037 & 0.1016 & 0.1093 \\
P(True) & 0.2556 & 0.2599 & 0.2559 & 0.2502 & 0.3067 & 0.2926 & 0.3123 & 0.3148 & 0.2492 & 0.2409 & 0.3753 & 0.3596 & 0.2626 & 0.3100 & 0.2626 & 0.2922 & 0.2054 & 0.2294 \\
\midrule
Probe & 0.5483 & 0.6382 & 0.6210 & 0.6908 & 0.4950 & 0.5296 & 0.4935 & 0.5563 & 0.5888 & 0.6487 & 0.4270 & 0.4511 & 0.4682 & 0.6084 & 0.5625 & 0.6763 & 0.4633 & 0.5779 \\
TPCs & 0.6032 & 0.6143 & 0.6063 & 0.6189 & 0.4304 & 0.4219 & 0.5138 & 0.5165 & 0.5740 & 0.5843 & 0.4338 & 0.4268 & 0.7726 & 0.7869 & 0.7054 & 0.7174 & 0.6228 & 0.6404 \\
ShieldHead & 0.3420 & 0.5392 & 0.3100 & 0.4990 & 0.2490 & 0.3721 & 0.0881 & 0.1271 & 0.0838 & 0.1535 & 0.1338 & 0.0942 & 0.1003 & 0.2415 & 0.0944 & 0.2142 & 0.1365 & 0.2204 \\
\midrule
Llama-Guard3 & 0.4007 & 0.4199 & 0.2535 & 0.2525 & 0.3650 & 0.3580 & 0.3774 & 0.3771 & 0.2542 & 0.2334 & 0.2923 & 0.2569 & 0.3670 & 0.4218 & 0.2226 & 0.2663 & 0.2648 & 0.2744 \\
ShieldGemma & 0.1521 & 0.1541 & 0.1404 & 0.1414 & 0.2897 & 0.2873 & 0.1659 & 0.1075 & 0.1568 & 0.1238 & 0.2633 & 0.2173 & 0.0232 & 0.0569 & 0.0591 & 0.0570 & 0.0812 & 0.0610 \\
Qwen3Guard & 0.2474 & 0.2660 & 0.1901 & 0.2071 & 0.2807 & 0.2226 & 0.2352 & 0.2099 & 0.1959 & 0.1814 & 0.2187 & 0.1381 & 0.1117 & 0.1898 & 0.1263 & 0.1705 & 0.1333 & 0.1393 \\
GPT-Safeguard & 0.2613 & 0.3503 & 0.1888 & 0.2391 & 0.2112 & 0.1187 & 0.2441 & 0.2555 & 0.1949 & 0.1930 & 0.2161 & 0.0997 & 0.2090 & 0.3608 & 0.1548 & 0.2597 & 0.1456 & 0.1977 \\
XGuard & 0.3128 & 0.3530 & 0.2520 & 0.2891 & 0.4891 & 0.5144 & 0.2810 & 0.2901 & 0.2386 & 0.2470 & 0.3723 & 0.3745 & 0.1311 & 0.2064 & 0.1819 & 0.2380 & 0.2738 & 0.3211 \\
PolyGuard & 0.5933 & 0.6214 & 0.5580 & 0.5913 & 0.5256 & 0.5369 & 0.5205 & 0.5402 & 0.5361 & 0.5512 & 0.4676 & 0.4670 & 0.6926 & 0.7584 & 0.6360 & 0.6887 & 0.6997 & 0.7449 \\
\midrule
Qwen3Guard-stream & 0.1151 & 0.0979 & 0.1172 & 0.0999 & 0.2538 & 0.2415 & 0.1919 & 0.1802 & 0.1617 & 0.1462 & 0.3162 & 0.3179 & 0.0101 & 0.0088 & 0.0391 & 0.0290 & 0.0697 & 0.0600 \\
SCM & 0.1747 & 0.1619 & 0.1556 & 0.1359 & 0.2526 & 0.2136 & 0.2014 & 0.1588 & 0.1698 & 0.1426 & 0.2699 & 0.2238 & 0.0186 & 0.0335 & 0.0602 & 0.0640 & 0.1017 & 0.0858 \\
\midrule
\textbf{ProbGuard-8B} & \textbf{0.0141} & \textbf{0.0249} & \textbf{0.0215} & \textbf{0.0259} & \textbf{0.0563} & \textbf{0.0539} & \textbf{0.0227} & \textbf{0.0275} & \textbf{0.0328} & \textbf{0.0466} & \textbf{0.0690} & \textbf{0.0989} & \textbf{0.0043} & \textbf{0.0070} & \textbf{0.0117} & \textbf{0.0130} & \textbf{0.0272} & \textbf{0.0347} \\
\bottomrule
\end{tabular}
\vspace{2pt}
\scriptsize
\raggedright
$^{*}$ Bold values indicate the best performance in each column.
\caption{Comparison of calibration performance between \textsc{ProbGuard} and 13 baselines across three target LLMs and three datasets at prefix length $k=10$.}
\label{tab:main_result}
\end{table*}

\subsection{Main Results}
\label{subsec:main_results}

\par\noindent\textbf{Calibration Performance.}
As shown in Table~\ref{tab:main_result}, \textsc{ProbGuard} consistently outperforms state-of-the-art methods, achieving superior Brier and ECE across nearly all datasets and target LLMs. 
For example, when targeting Qwen3-8B on the PKU dataset, \textsc{ProbGuard} achieves a Brier and ECE of 0.0141 and 0.0249, respectively.
In comparison, the strongest baseline in this category (Qwen3Guard-stream) attains a Brier of 0.1151 and an ECE of 0.0979.
Notably, \textsc{ProbGuard} demonstrates a distinct advantage in calibration performance across different target LLMs.
Specifically, \textsc{ProbGuard} reduces the average Brier score by approximately 91\%, 81\%, and 85\% relative to Llama-Guard3, ShieldGemma, and Qwen3Guard, respectively.

The superior calibration performance of \textsc{ProbGuard} stems from its direct use of LLM output distributional signals to estimate calibrated safety risk, thereby preserving information about the uncertainty over possible continuations before the full response is observed while avoiding dependence on LLM hidden states.
In contrast, confidence-based methods (e.g., Verbalized and P(True)) hinge on the LLM's own risk estimation for each input, which inevitably introduces instability into their outputs.
Although guardrails (e.g., Llama-Guard3 and ShieldGemma) classify discrete token sequences into deterministic safety labels, they cannot estimate the safety risk over possible continuations from the current generation state.
Streaming monitors (e.g., SCM and Qwen3Guard-stream) assess intermediate generation states using information derived from the observed prefix, which characterizes only the realized generation trajectory rather than the uncertainty over possible continuations.
Feature-probing methods (e.g., Probe and TPCs) assess safety from internal hidden states, whose architecture-specific representations prevent the resulting monitor from being applied across different LLMs.


\par\noindent\textbf{Early Intervention against Jailbreak Attacks.}
As shown in Table~\ref{tab:baseline_attack_advbench}, \textsc{ProbGuard} consistently outperforms state-of-the-art methods, achieving the best defense performance across nearly all jailbreak attacks using only the first 10 decoding steps.
For example, \textsc{ProbGuard-8B} achieves an overall average ASR of 0.75\% across both datasets, while the best baseline (Llama-Guard3) achieves 2.42\%.
Furthermore, \textsc{ProbGuard} with smaller model sizes also demonstrates competitive defense effectiveness: \textsc{ProbGuard-4B} attains an overall average ASR of 1.00\%, still surpassing the best baseline, and \textsc{ProbGuard-0.6B} achieves 2.83\%.
These results confirm that early output distributional signals alone are sufficient for ProbGuard to reliably estimate safety risks and intervene against jailbreak attacks in time.

\subsection{Further Analysis}
\label{subsec:further_analysis}

\begin{table*}[t]
\centering
\fontsize{7}{8}\selectfont
\begin{threeparttable}
\setlength{\tabcolsep}{4pt}
\renewcommand{\arraystretch}{1.12}
\begin{tabular}{l c c c c c c c c c c c c c c c}
\toprule
Method & \multicolumn{6}{c}{AdvBench} & Avg. & \multicolumn{6}{c}{HarmBench} & Avg. \\
\cmidrule(lr){2-7} \cmidrule(lr){9-14}
& GCG & COLD & AdvPrefix & AdvPrompter & PAIR & Eclipse & & GCG & COLD & AdvPrefix & AdvPrompter & PAIR & Eclipse & \\
\midrule
No defense & 54.0\% & 47.0\% & 55.0\% & 62.0\% & 58.0\% & 47.0\% & 53.83\% & 31.0\% & 32.0\% & 24.0\% & 43.0\% & 57.0\% & 40.0\% & 37.83\% \\
\midrule
TPCs & 17.0\% & 4.0\% & 25.0\% & 7.0\% & 1.0\% & 1.0\% & 9.17\% & 2.0\% & 2.0\% & 1.0\% & 3.0\% & 5.0\% & 2.0\% & 2.50\% \\
ShieldHead & 2.0\% & 2.0\% & 7.0\% & 21.0\% & 12.0\% & 0\% & 7.33\% & 4.0\% & 4.0\% & 3.0\% & 9.0\% & 10.0\% & 3.0\% & 5.50\% \\
\midrule
Llama-Guard3 & 2.0\% & 1.0\% & 2.0\% & 9.0\% & 6.0\% & 2.0\% & 3.67\% & 1.0\% & 2.0\% & 0\% & 3.0\% & 0\% & 1.0\% & 1.17\% \\
ShieldGemma & 17.0\% & 5.0\% & 26.0\% & 4.0\% & 9.0\% & 4.0\% & 10.83\% & 2.0\% & 3.0\% & 0\% & 4.0\% & 10.0\% & 7.0\% & 4.33\% \\
Qwen3Guard-8B & 15.0\% & 1.0\% & 14.0\% & 6.0\% & 1.0\% & 2.0\% & 6.50\% & 3.0\% & 2.0\% & 1.0\% & 0\% & 2.0\% & 0\% & 1.33\% \\
GPT-Safeguard & 9.0\% & 3.0\% & 13.0\% & 5.0\% & 4.0\% & 6.0\% & 6.67\% & 7.0\% & 4.0\% & 1.0\% & 8.0\% & 15.0\% & 6.0\% & 6.83\% \\
XGuard & 14.0\% & 0\% & 19.0\% & 6.0\% & 2.0\% & 0\% & 6.83\% & 2.0\% & 1.0\% & 1.0\% & 0\% & 0\% & 0\% & 0.67\% \\
PolyGuard & 7.0\% & 2.0\% & 5.0\% & 6.0\% & 3.0\% & 0\% & 3.83\% & 7.0\% & 4.0\% & 1.0\% & 4.0\% & 6.0\% & 4.0\% & 4.33\% \\
\midrule
Qwen3Guard-stream & 30.0\% & 7.0\% & 40.0\% & 11.0\% & 4.0\% & 8.0\% & 16.67\% & 3.0\% & 1.0\% & 1.0\% & 1.0\% & 3.0\% & 4.0\% & 2.17\% \\
SCM & 3.0\% & 2.0\% & 4.0\% & 13.0\% & 7.0\% & 1.0\% & 5.00\% & 4.0\% & 4.0\% & 1.0\% & 5.0\% & 11.0\% & 1.0\% & 4.33\% \\
\midrule
\textbf{ProbGuard-8B} & 0\% & 1.0\% & 1.0\% & 1.0\% & 1.0\% & 1.0\% & 0.83\% & 0\% & 1.0\% & 0\% & 1.0\% & 1.0\% & 1.0\% & 0.67\% \\
\textbf{ProbGuard-4B} & 1.0\% & 1.0\% & 0\% & 1.0\% & 2.0\% & 0\% & 0.83\% & 1.0\% & 0\% & 0\% & 2.0\% & 3.0\% & 1.0\% & 1.17\% \\
\textbf{ProbGuard-0.6B} & 2.0\% & 3.0\% & 4.0\% & 1.0\% & 6.0\% & 1.0\% & 2.83\% & 2.0\% & 1.0\% & 0\% & 5.0\% & 7.0\% & 2.0\% & 2.83\% \\
\bottomrule
\end{tabular}
\begin{tablenotes}
\fontsize{7}{9}\selectfont
   \item \hspace{-1.2em}$^{*}$ Operating thresholds for all defense methods are determined by maximizing the F1 score on the held-out PKU validation split.
\end{tablenotes}
\end{threeparttable}
\caption{Comparison of ASRs for Qwen3-8B on AdvBench and HarmBench under \textsc{ProbGuard} and baselines at a prefix length $k=10$ (\%). All ASRs are evaluated by GPT-5 as the safety judge for generated responses.}
\label{tab:baseline_attack_advbench}
\end{table*}

\par
\begin{figure}[h]
    \centering
    \includegraphics[width=1.0\columnwidth]{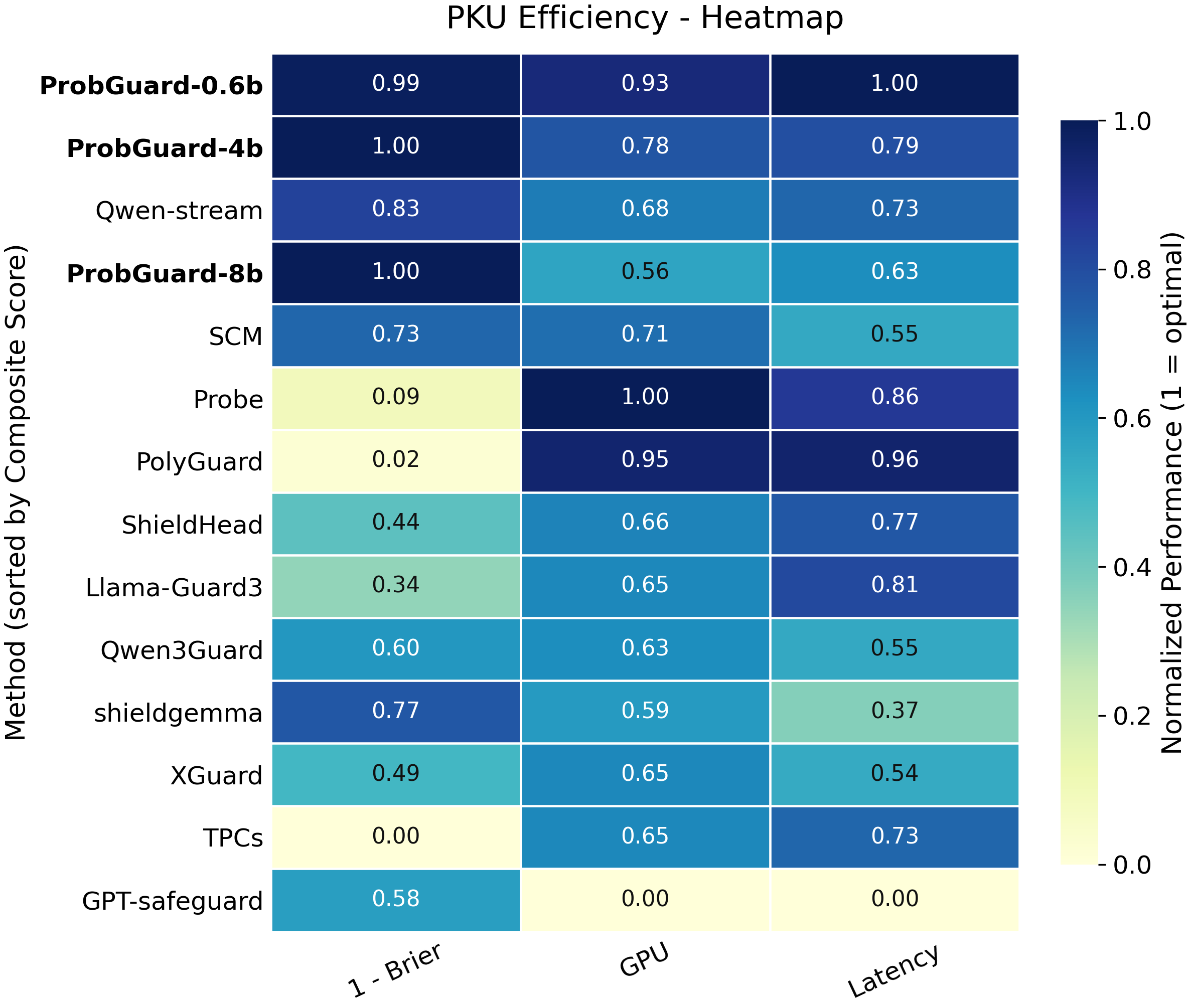}
    \caption{Comparison of normalized efficiency scores between \textsc{ProbGuard} and baselines on PKU with higher scores indicating better performance.}
    \label{fig:efficiency_heatmap}
\end{figure}

\noindent\textbf{Efficiency Analysis.}
As shown in Figure~\ref{fig:efficiency_heatmap}, \textsc{ProbGuard} balances calibration performance with computational cost. The heatmap reports column-wise min--max normalized values for $1-\mathrm{Brier}$, GPU memory, and latency, where higher values indicate better performance. For GPU memory and latency, the normalization is inverted so that lower raw costs receive higher scores. Raw measurements are reported in the supplementary material.
At prefix length $k=10$, \textsc{ProbGuard-8B} estimates calibrated safety risk directly from output distributional signals, without waiting for a complete target response or generating a textual defense decision. It processes 1,000 samples in 36.4~s, reducing latency by approximately 52.7\% compared with GPT-Safeguard (76.9s), while achieving the best calibration performance ($1-\mathrm{Brier}=0.9859$).
For lower computational budgets, \textsc{ProbGuard-0.6B} reduces latency to 12.8~s and GPU memory to 3.32~GB while maintaining a strong calibration performance of 0.9782.

\begin{figure}
    \centering
    \includegraphics[width=1.0\linewidth]{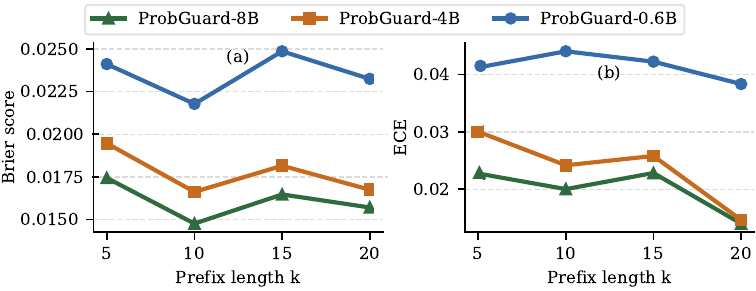}
\caption{Comparison of calibration performance across prefix lengths for three \textsc{ProbGuard} model scales. Lower Brier and ECE indicate better calibrated risk estimation.}
\label{fig:prefix_length}
\end{figure}

\par\noindent\textbf{Calibration across Decoding Steps.}
As shown in Figure~\ref{fig:prefix_length}, we evaluate the calibration performance of \textsc{ProbGuard} at multiple prefix lengths using Brier score and ECE, where lower values indicate more accurate risk estimation. As additional output distributional signals are observed, all three \textsc{ProbGuard} models show an overall reduction in calibration error. From $k=5$ to $k=20$, including the extrapolated evaluation point beyond the training range, the Brier scores decrease from 0.0174 to 0.0157, from 0.0195 to 0.0167, and from 0.0241 to 0.0232 for \textsc{ProbGuard-8B}, \textsc{ProbGuard-4B}, and \textsc{ProbGuard-0.6B}, respectively, while their ECE values decrease from 0.0227 to 0.0140, from 0.0300 to 0.0146, and from 0.0412 to 0.0383. These results demonstrate that \textsc{ProbGuard} is not calibrated only at a particular prefix position, but continuously integrates the accumulated output distributional signals to refine its risk predictions throughout generation. Despite minor fluctuations at intermediate prefix lengths, the consistent endpoint improvements across model scales confirm that additional output distributional signals generally enable more accurate and reliable calibration.

\par\noindent\textbf{Effect of Monte Carlo Sampling Budget.}
As shown in Table~\ref{tab:n_selection}, we evaluate the impact of sampling budget on the quality of calibration target estimation, using the results with $N=128$ as the reference.
We quantify estimation stability using RMSE and Spearman correlation, which measure the accuracy and rank-order consistency of the estimated safety probabilities, as well as Flip Rate, which we define as the fraction of samples whose thresholded safe/unsafe decision changes relative to the $N=128$ reference.
The improvement is substantial up to $N=16$, which achieves a Flip Rate of 3.60\% and a Spearman correlation of 0.9197, but the marginal gain diminishes thereafter.
Increasing to $N=32$ reduces RMSE by only 0.0251 while doubling the sampling cost.
We therefore adopt $N=16$ as the default budget, as it provides a favorable balance between estimation accuracy and computational efficiency.

\begin{table}[t]
\centering
\begin{threeparttable}
\setlength{\tabcolsep}{5pt}
\renewcommand{\arraystretch}{1.2}
\fontsize{8}{9}\selectfont
\begin{tabular*}{\columnwidth}{@{\extracolsep{\fill}}lccc@{}}
\toprule
\textbf{Sampling Budget} & \textbf{RMSE $\downarrow$} & \textbf{Spearman $\uparrow$} & \textbf{Flip Rate $\downarrow$} \\
\midrule
$N=8$   & 0.1014 & 0.8557 & 5.60\% \\
\textbf{$N=16$} & 0.0711 & 0.9197 & 3.60\% \\
$N=32$  & 0.0460 & 0.9597 & 2.80\% \\
$N=64$  & 0.0254 & 0.9879 & 0.80\% \\
$N=128$ & 0.0000 & 1.0000 & 0.00\% \\
\bottomrule
\end{tabular*}
\begin{tablenotes}
\fontsize{7}{8}\selectfont
\item $^{*}$ Metrics are computed against the $N=128$ reference, and bold indicates the default sampling budget. Flip Rate is the percentage of samples whose thresholded safe/unsafe decision differs from the $N=128$ reference.
\end{tablenotes}
\end{threeparttable}
\caption{Comparison of Monte Carlo sampling budgets for Qwen3-8B on PKU at a prefix length of $k=10$.}
\label{tab:n_selection}
\end{table}

\begin{table}[t]
\vspace{-0.5em}
\centering
\setlength{\tabcolsep}{3pt}
\renewcommand{\arraystretch}{1.2}
\fontsize{8}{9}\selectfont
\begin{tabular*}{\columnwidth}{@{\extracolsep{\fill}}lcccccc@{}}
\toprule
\multirow{2}{*}{\textbf{Method}} &
\multicolumn{2}{c}{\textbf{Qwen3-8B}} &
\multicolumn{2}{c}{\textbf{Llama3-8B-it}} &
\multicolumn{2}{c}{\textbf{Gemma2-9B-it}} \\
\cmidrule(lr){2-3} \cmidrule(lr){4-5} \cmidrule(lr){6-7}
& \textbf{Brier} & \textbf{ECE}
& \textbf{Brier} & \textbf{ECE}
& \textbf{Brier} & \textbf{ECE} \\
\midrule
Token & 0.0159 & 0.0179 & 0.0286 & 0.0409 & 0.0055 & 0.0058 \\
Hidden state$^{*}$     & 0.0886 & 0.0290 & $\times$ & $\times$ & $\times$ & $\times$ \\
\textbf{ProbGuard-8B}  & 0.0141 & 0.0249 & 0.0227 & 0.0275 & 0.0043 & 0.0070 \\
\bottomrule
\end{tabular*}
\begin{tablenotes}
\fontsize{7}{8}\selectfont
\item $^{*}$ The hidden-state probe trained solely on Qwen3-8B does not transfer to other LLMs.
\end{tablenotes}
\caption{Comparison of calibration performance across different input representations on PKU with Qwen3-8B training.}
\label{tab:ablation_goal}
\end{table}

\par\noindent\textbf{Calibration Performance with Different Input Representations.}
As shown in Table~\ref{tab:ablation_goal}, we compare three input representations for training \textsc{ProbGuard} under a fixed calibration target on the PKU dataset: token-based probabilities, hidden-state activations, and output distributional signals as our default.
Token-based representations achieve competitive calibration on Qwen3-8B (Brier/ECE: 0.0159/0.0179) and Gemma2-9B-it (0.0055/0.0058), but their performance deteriorates on Llama3-8B-it (0.0286/0.0409), indicating less consistent transfer across target LLMs.
Hidden-state activations yield worse results on Qwen3-8B (0.0886/0.0290) and are inapplicable to other architectures.
In contrast, output distributional signals consistently achieve strong calibration across all three target models, with Brier/ECE of 0.0141/0.0249, 0.0227/0.0275, and 0.0043/0.0070 on Qwen3-8B, Llama3-8B-it, and Gemma2-9B-it, respectively.
These results demonstrate that output distributional signals, as an architecture-agnostic representation, enable robust safety risk estimation across different LLM families.


\par\noindent\textbf{Evaluation of \textsc{CalibEval}.}
As shown in Table~\ref{tab:local_judge}, \textsc{CalibEval} consistently outperforms LLM-based judges, achieving superior F1 scores with the lowest time cost. 
For example, when evaluated on the PKU dataset, \textsc{CalibEval} achieves an F1 of 0.943 within just 4.9s, significantly outperforming the best LLM-based judge (Gemini3-pro), which yields only 0.833 at the cost of 6009s. 
Notably, \textsc{CalibEval} also attains an extremely low FPR of 0.058, far below those of all three LLM-based judges, which is critical for avoiding over‑penalizing benign responses. 
These results confirm that \textsc{CalibEval} provides accurate and efficient safety labels for the sampled continuations, enabling reliable calibration target construction in ProbGuard.


\begin{table}[t]
\centering
\setlength{\tabcolsep}{5pt}
\renewcommand{\arraystretch}{1.2}
\fontsize{8}{9}\selectfont
\begin{threeparttable}
\begin{tabular*}{\columnwidth}{@{\extracolsep{\fill}}lccccc@{}}
\toprule
Judges & Acc & F1$^{\ddagger}$ & TPR & FPR & Time(s)$^*$ \\
\midrule
DeepSeek-R1 & 0.737 & 0.786 & 0.964 & 0.490 & 2321 \\
GPT-5 & 0.791 & 0.827 & 0.996 & 0.414 & 2148 \\
Gemini3-Pro & 0.801 & 0.833 & 0.994 & 0.392 & 6009 \\
\textbf{\textsc{CalibEval} (Ours)} & \textbf{0.943} & \textbf{0.943} & 0.944 & \textbf{0.058} & \textbf{4.9} \\
\bottomrule
\end{tabular*}
\begin{tablenotes}
\fontsize{7}{8}\selectfont
\item $^*$ Time denotes total evaluation time for 1,000 responses.
\item $^{\ddagger}$ The ground truth labels of the eval dataset are from the dataset's own annotations.
\end{tablenotes}
\end{threeparttable}
\caption{Comparison of judge-model effectiveness for calibration-target construction on the PKU evaluation set.}
\label{tab:local_judge}
\end{table}

\section{Conclusion}
\label{sec:conclusion}
In this work, we propose \textsc{ProbGuard}, the first completely probabilistic, architecture-agnostic guardrail to calibrate safety risk based on LLM early output distributional signals. We formulate the safety risk as the unsafe probability of continued generation and employ Monte Carlo sampling for estimation. We further derive unified probability-weighted representations for different LLM output distributions, allowing \textsc{ProbGuard} to learn for safety risk calibration on distributional signals. Extensive experimental results demonstrate that \textsc{ProbGuard} achieves superior calibration performance across different LLMs and safety datasets, enabling effective early stopping of ongoing unsafe outputs induced by different jailbreak attacks.

\bibliographystyle{aaai2027}
\bibliography{aaai2027}

\end{document}